\pdfoutput=1
\documentclass[11pt]{article}

\usepackage{ACL2023}

\usepackage{times}
\usepackage{latexsym}

\usepackage[T1]{fontenc}
\usepackage[utf8]{inputenc}
\usepackage[main=english,ukrainian]{babel}

\usepackage{microtype}

\usepackage{inconsolata}
\usepackage{booktabs}
\usepackage{graphicx}
\usepackage{color, colortbl}
\usepackage{enumitem}
\usepackage{caption}
\usepackage{adjustbox}
\usepackage{amssymb}
\usepackage{utfsym}
\usepackage{mathtools}
\usepackage{amsmath}
\usepackage{subcaption}

\usepackage{xcolor}

\definecolor{Gray}{gray}{0.9}

\usepackage{xcolor}

\newcommand{\greencheck}{{\color{green}\usym{2714}}}

\newcommand{\redcross}{{\color{red}\usym{2718}}}

\title{Toxicity Classification in Ukrainian}

\author{Daryna Dementieva\textsuperscript{1}, Valeriia Khylenko\textsuperscript{1}, Nikolay Babakov\textsuperscript{2} \and {\bf Georg Groh}\textsuperscript{1} \\
\textsuperscript{1} TU Munich, Department of Informatics, Germany\\ 
\textsuperscript{2}Centro Singular de Investigación en Tecnoloxías Intelixentes (CiTIUS), \\ Universidade de Santiago de Compostela \\
\href{mailto:daryna.dementieva@tum.de}{\texttt{\small daryna.dementieva@tum.de}},
\href{mailto:nikolay.babakov@usc.es}{\texttt{\small nikolay.babakov@usc.es}},
\href{mailto:a.panchenko@skol.tech}{\texttt{\small grohg@in.tum.de}}
}

\begin{document}
\maketitle
\begin{abstract}
The task of toxicity detection is still a relevant task, especially in the context of safe and fair LMs development.  Nevertheless, labeled binary toxicity classification corpora are not available for all languages, which is understandable given the resource-intensive nature of the annotation process. Ukrainian, in particular, is among the languages lacking such resources.
To our knowledge, there has been no existing toxicity classification corpus in Ukrainian. In this study, we aim to fill this gap by investigating cross-lingual knowledge transfer techniques and creating labeled corpora by: (i)~translating from an English corpus, (ii)~filtering toxic samples using keywords, and (iii)~annotating with crowdsourcing. We compare LLMs prompting and other cross-lingual transfer approaches with and without fine-tuning offering insights into the most robust and efficient baselines. \\ \textbf{\textcolor{red}{\textit{This paper contains rude texts that only serve as illustrative examples.}}}
\end{abstract}

\section{Introduction}
%more and more multilingual model, huge improvement of translation systems, adapters, emerging abilities of LLMs, but yet not tested on typical NLP tasks for other languages rather then English
% We want to address this gap
% three tasks never explored for Ukrainian before, four approaches, new data (even if translated), additionally, natural test sets for the tasks.

Lately, the NLP community has shifted away from exclusively developing monolingual English models and is placing greater emphasis on the development of fair multilingual NLP technologies. There were released plenty of multilingual models, i.e. mBERT~\cite{devlin-etal-2019-bert}, XLM-RoBERTa~\cite{conneau2019unsupervised}, mT5~\cite{xue2020mt5}, mBART~\cite{tang2020multilingual}, BLOOMz~\cite{muennighoff2022crosslingual}, NLLB~\cite{costa2022no}. Additionally, Large Language Models (LLMs) pre-trained on extensive corpora have expanded the realm of potential capabilities~\cite{wei2022emergent} not only for novel tasks but also for languages.

\begin{table}[t!]
    \centering
    \footnotesize
    \begin{tabular}{p{1.70cm}|p{5.3cm}}
    \toprule 
        Toxic & \foreignlanguage{ukrainian}{І ніх*шеньки їй за те не буде.} \newline \textcolor{gray}{\scriptsize{\textit{And she's not going to get a f*king thing for it.}}} \\ 
         & \foreignlanguage{ukrainian}{А зі всіх компліментів які мені казали, це те що я п*ар} \newline \textcolor{gray}{\scriptsize{\textit{And of all the compliments I've been given, the only one I've received is that I'm a f*got.}}} \\ 
          & \foreignlanguage{ukrainian}{Увесь твіттер у ваших *бучих котах.} \newline \textcolor{gray}{\scriptsize{\textit{The whole of Twitter is in your f*king cats.}}} \\ 
        \midrule
        Non-toxic & \foreignlanguage{ukrainian}{І знову дві години на прокидання.} \newline \textcolor{gray}{\scriptsize{\textit{And again, two hours to wake up.}}} \\
         & \foreignlanguage{ukrainian}{Ну, це тіпа добре, коли хвалять.} \newline \textcolor{gray}{\scriptsize{\textit{Well, it's kind of nice to be praised.}}} \\
         & \foreignlanguage{ukrainian}{скоро буду своєю серед чужих))) аха} \newline \textcolor{gray}{\scriptsize{\textit{soon I will be my own among strangers))) aha}}} \\
        \bottomrule
    \end{tabular}
    \caption{Toxic and non-toxic examples in Ukrainian.}
    \label{tab:intro_examples}
\end{table}

\begin{table*}[ht!]
    \centering
    \footnotesize
    \begin{tabular}{p{1.8cm}|p{3.2cm}|p{3.2cm}|c|c|c|c}
    \toprule
        \textbf{Method} & \textbf{Models} & \textbf{Datasets} & \textbf{\shortstack{Translation \\ Dependence}} & \textbf{\shortstack{Data \\ Creation}} & \textbf{\shortstack{Fine \\ tuning}} & \textbf{\shortstack{\# Inference \\ Steps }} \\
        \midrule
        \multicolumn{6}{c}{\textbf{\textit{Cross-lingual Transfer Methods}}} \\
        \midrule
        \textit{Backtranslation} & - Toxicity detection model for the resource-rich language; \newline - Translation model from resource-rich to the target language; & \multicolumn{1}{c|}{---} & \greencheck & \redcross & \redcross & 3 \\
        \hline
        \textit{LLM \newline prompting} & - LLM with the knowledge of the resource-rich language and (emerging) knowledge of the target language; & \multicolumn{1}{c|}{---} & \redcross & \redcross & \redcross & 1 \\
        \hline
        \textit{Adapter \newline Training} & - Auto-regressive multilingual LM where the resource-rich and target languages are present; - Language adapter layers for both languages; & - Toxicity classification dataset in the resource-rich language; \newline - Corpus for translation between the resource-rich and target languages; & \redcross & \redcross & \greencheck & 1 \\
        \midrule
        \multicolumn{6}{c}{\textbf{\textit{Data Acquisition Methods}}} \\
        \midrule
        \textit{Training Data Translation} & - Translation model to the target language; \newline - Auto-regressive multilingual or monolingual LM for the target language; & - Toxicity classification dataset in the resource-rich language; & \greencheck & \greencheck & \greencheck & 1 \\
        \hline
        \textit{Semi-synthetic data by keywords filtering} & - Embedding model of texts in the target language; & - Texts in the target language; \newline - List of toxic keywords in the target language; & \redcross & \greencheck & \greencheck & 1 \\
        \hline
        \textit{Crowdsourcing data filtering} & - Embedding model of texts in the target language; & - Texts in the target language; & \redcross & \greencheck & \greencheck & 1 \\
    \bottomrule
    \end{tabular}
    \caption{Comparison of the considered approaches for cross-lingual detoxification transfer and corpora acquisition based on required computational and data resources.}
    \label{tab:resources_approaches}
\end{table*}

Nevertheless, the coverage of languages and classical NLP tasks corpora existence is still unequal. In the scope of harmful language detection, we discovered an absence of any toxicity or hateful detection corpora for the Ukrainian language. Thus, the question arises: what is the most effective and promising approach to acquiring a binary toxicity classification corpus for a new language, considering all the recent advancements in the field of NLP. Answering this main research question, the contribution of this work are the following:
\begin{itemize}
    \item We present the first of its kind toxicity classification corpus for Ukrainian (Table~\ref{tab:intro_examples}) testing three approach for its acquisition: (i) translation from a resource rich language; (ii)~toxic samples filtering by toxic keywords; (iii)~crowdsourcing data annotation;
    \item Additionally, we explore three types of cross-lingual knowledge transfer approaches---Backtranslation, LLMs Prompting, and Adapter Training; 
    \item We test both cross-lingual and supervised approaches on all test sets providing insights into the methods effectiveness.
\end{itemize}
All the obtained data and models are available for the public usage online.\footnote{\scriptsize{\href{https://huggingface.co/ukr-detect}{https://huggingface.co/ukr-detect}}}\textsuperscript{,}\footnote{\scriptsize{\href{https://huggingface.co/textdetox}{https://huggingface.co/textdetox}}}\textsuperscript{,}\footnote{\scriptsize{\href{https://huggingface.co/dardem/xlm-roberta-large-uk-toxicity}{https://huggingface.co/dardem/xlm-roberta-large-uk-toxicity}}}

\section{Related Work}
% couple of cross-lingual transfer methods
% list couple multilingual corpora for the tasks. BUT THERE IS NO UKRAINIAN THERE.

% The topic of cross-lingual knowledge transfer appeared recently but has already been studied across multiple languages.
The usual case for cross-lingual transfer setup is when data for a specific task is available for English but none for the target language.
In such a setup, translation of training data approach has been already explored for sentiment analysis~\cite{DBLP:journals/apin/KumarPR23} and offensive texts classification \cite{el2022multilingual,DBLP:journals/csse/WadudMSNS23}.

For toxicity, both monolingual and multilingual corpora have been introduced. Thus, English Jigsaw dataset~\cite{jigsaw} was later extended to the multilingual format~\cite{jigsaw_multilingual}. Within East European language, there were presented offensive language detection in Polish~\cite{DBLP:journals/data/PtaszynskiPDSSF24} and Serbian~\cite{DBLP:conf/ldk/JokicSKS21} based on Twitter data. In the related domain, Ukrainian bullying detection system was developed based on translated English data in \cite{oliinyk2023low}. However, none of the works yet covered specifically Ukrainian toxicity detection. 

\paragraph{Definition of Toxicity} While there can be different types of toxic language in conversations~\cite{DBLP:conf/acl-alw/PriceGFMRSTDS20,DBLP:conf/euspn/GildaGS021}, i.e. sarcasm, hate speech, direct insults, in this work include samples with substrings that are commonly referred to as vulgar or profane language~\cite{costa2022no,logacheva-etal-2022-paradetox} while the whole main message can be both neutral and toxic. Thus, we are considering the task of binary toxicity classification assigning the labels either toxic or non-toxic.

% For the text classification tasks covered in this work, both monolingual and multilingual datasets were constructed. ... However, none of them has covered the Ukrainian language. 

\section{Cross-lingual Knowledge Transfer Methods}

\begin{table*}[ht!]
\centering
\footnotesize
\begin{tabular}{@{}llll@{}}
\toprule
& Translated dataset & Semi-synthetic dataset & Crowdsourced dataset \\
\midrule
\multicolumn{1}{l|}{Train}   
& \begin{tabular}[c]{@{}l@{}}total: 24616\\ toxic: 12307\\ non-toxic: 12309\end{tabular} 
& \begin{tabular}[c]{@{}l@{}}total: 12606\\ toxic: 6362\\ non-toxic: 6244\end{tabular} 
& \begin{tabular}[c]{@{}l@{}}total: 3000\\ toxic: 1500\\ non-toxic: 1500\end{tabular} \\ 
\midrule
\multicolumn{1}{l|}{Val}          
& \begin{tabular}[c]{@{}l@{}}total: 4000\\ toxic: 2000\\ non-toxic: 2000\end{tabular}    
& \begin{tabular}[c]{@{}l@{}}total: 4202\\ toxic: 2071\\ non-toxic: 2131\end{tabular}      
& \begin{tabular}[c]{@{}l@{}}total: 1000\\ toxic: 500\\ non-toxic: 500\end{tabular} \\ 
\midrule
\multicolumn{1}{l|}{Test}         
& \begin{tabular}[c]{@{}l@{}}total: 52294\\ toxic: 5800\\ non-toxic: 46494\end{tabular}  
& \begin{tabular}[c]{@{}l@{}}total: 4214\\ toxic: 2114\\ non-toxic: 2008\end{tabular}       
& \begin{tabular}[c]{@{}l@{}}total: 1000\\ toxic: 500\\ non-toxic: 500\end{tabular} \\ 
\bottomrule
% \multicolumn{1}{l|}{Natural test} & \begin{tabular}[c]{@{}l@{}}total: 4214\\ toxic: 2114\\ non-toxic: 2088\end{tabular}    & \begin{tabular}[c]{@{}l@{}}total: 3000\\ formal: 1500\\ informal: 1500\end{tabular}       & \begin{tabular}[c]{@{}l@{}}total: 901\\ neutral: 300\\ contradiction: 300\\ entailment: 301\end{tabular}             \\ \bottomrule
\end{tabular}
\caption{Statistics of the obtained datasets: train/val/test splits.}
\label{tab:dataset_stat}
\end{table*}

Firstly, we test three cross-lingual knowledge transfer methods that do not require any training data in the target language acquisition (Table~\ref{tab:resources_approaches}): (i)~Backtranslation; (ii)~LLM Prompting; (iii)~Adapter Training. We assume a setup where resource-rich available language is English.

\paragraph{Backtranslation} For many tasks, an English classifier may already exist, making it a natural baseline to translate the input text from Ukrainian to English and then employ the English classifier for the task. This Backtranslation approach eliminates the need for fine-tuning but relies on external models—an translation system and an English classifier— for consistent functionality.

\paragraph{LLM Prompting} The next approach that as well does not require fine-tuning is prompting of LLMs. Current advances in generative models showed the feasibility of transforming any NLP classification task into text generation task \cite{chung2022scaling,DBLP:conf/emnlp/AlySLZW23}. Thus, the prompt can be designed in a zero-shot or a few-shot manner requesting the model to answer with the label. While LLMs were already tested for a hate speech classification task for multiple languages~\cite{das2023evaluating}, there were no yet experiments for any text classification task for Ukrainian language which might be underrepresented in such models. We provide the final design of our prompt in Appendix~\ref{sec:app_llm_prompts}.

\paragraph{Adapter Training} Finally, the most parameter-efficient approach involves employing language-specific Adapter layers~\cite{pfeiffer-etal-2020-adapterhub}. Such a layer, firstly, for English, can be added upon multilingual LM. Everything remains frozen while fine-tuning of the final Adapter for the downstream task. Then, English Adapter is replaced with Ukrainian one and inference for the task in the target language can be performed. 

\section{Data Acquisition Methods}
To obtain supervised detection models, we test three ways of training data acquisition for toxicity detection task (Table~\ref{tab:resources_approaches}): (i) English toxicity corpus translation into Ukrainian; (ii)~filtering toxic samples by pre-defined dictionary of Ukrainian toxic keywords; (iii) crowdsourcing annotation to filter Twitter corpus into toxic and non-toxic samples. The examples of samples from these three dataset can be found in Appendix~\ref{sec:app_data}.

\subsection{Training Corpus Translation} 

To avoid the permanent dependence on a translation system per each request, we can translate the whole English dataset and, as a result, get synthetic training data for the task. Then, a downstream task fine-tuning is possible. This approach's main advantage is that there are no external dependencies during the inference time, but it requires computational resources for fine-tuning. Moreover, some class information might vanish after translation and will not be adapted for the target language. 

\paragraph{English Dataset} To test this approach, we considered English datasets Jigsaw data~\cite{jigsaw}. We collapsed all labels except from ``non-toxic'' into one ``toxic'' class.

\paragraph{Translation Systems Choice} To choose the most appropriate translation system, we took into consideration two opensource models---NLLB\footnote{\href{https://huggingface.co/facebook/nllb-200-distilled-600M}{https://huggingface.co/facebook/nllb-200-distilled-600M}}~\cite{costa2022no} and Opus\footnote{\href{https://huggingface.co/Helsinki-NLP/opus-mt-en-uk}{https://huggingface.co/Helsinki-NLP/opus-mt-en-uk}}~\cite{tiedemann-2012-parallel}. We randomly selected 50 samples per each dataset and asked 3 annotators (native speakers in Ukrainian) to verify the quality. As a result, we choose Opus translation system for toxicity classification as it preserves better the toxic lexicon. The system achieved 90\% of qualitative translations.

\begingroup
\renewcommand{\arraystretch}{1.05}
\begin{table*}[ht!]
\footnotesize
\centering

\begin{tabular}{p{4.25cm}|ccc|ccc|ccc}
\toprule

& \textbf{Pr} & \textbf{Re} & \textbf{F1} & \textbf{Pr} & \textbf{Re} & \textbf{F1} & \textbf{Pr} & \textbf{Re} & \textbf{F1} \\ \midrule

& \multicolumn{3}{c|}{{Translated Test Set}} & \multicolumn{3}{c|}{{Semi-synthetic Test Set}} & \multicolumn{3}{c}{{Crowdsourced Test Set}} \\ \midrule
\multicolumn{10}{c}{\textit{Prompting of LLMs}} \\
\midrule

LLaMa-2 Prompting & 0.50 & 0.67 & 0.42 & 0.67 & 0.49 & 0.67 & 0.24 & 0.50 & 0.32  \\
Mistral Prompting &
  \textbf{0.68} &
  \textbf{0.74} &
  \textbf{0.70} &
  \textbf{0.81} &
  \textbf{0.76} &
  \textbf{0.75} & \textbf{0.56} & \textbf{0.68} & \textbf{0.52} \\
\midrule
\multicolumn{10}{c}{\textit{Cross-lingual transfer approaches}} \\
\midrule
Backtranslation & \multicolumn{3}{c|}{---}  & \textbf{0.76} & \textbf{0.56} & \textbf{0.58} & \textbf{0.75} & \textbf{0.68} & \textbf{0.65} \\
Adapter Training & 0.66 & 0.63 & 0.65 & 0.66 & 0.58 & 0.52 & 0.64 & 0.58 & 0.53 \\
\midrule
\multicolumn{10}{c}{\textit{Fine-tuning of LMs on different types of data}} \\
\midrule
XLM-R-finetuned-translated & \textcolor{gray}{0.68} & \textcolor{gray}{0.86} & \textcolor{gray}{0.70} & 0.79 & 0.77 & 0.77 & 0.70 & \textbf{0.68} & \textbf{0.67}\\
XLM-R-finetuned-semisynthetic &  0.59 & 0.53 & 0.53 & \textcolor{gray}{0.99} & \textcolor{gray}{0.99} & \textcolor{gray}{0.99} & \textbf{0.75} & 0.57 & 0.48\\
XLM-R-finetuned-crowdsourced & \textbf{0.61} & \textbf{0.63} & \textbf{0.62} & \textbf{0.93} & \textbf{0.93} & \textbf{0.93} & \textcolor{gray}{0.99} & \textcolor{gray}{0.99} & \textcolor{gray}{0.99} \\

\bottomrule
\end{tabular}
\caption{Ukrainian Toxicity Classification results. Within methods comparison, \textbf{bold} numbers denote the best results within methods types, \textcolor{gray}{gray}---in domain results of the fine-tuned models. We do not test Backtranslation approach on the translated data as we cannot guarantee this test set was not present in the English training data of the model.} 
\label{tab:all_results}
\end{table*}
\endgroup

\begin{figure}[h!]
\centering
\includegraphics[scale=0.8]{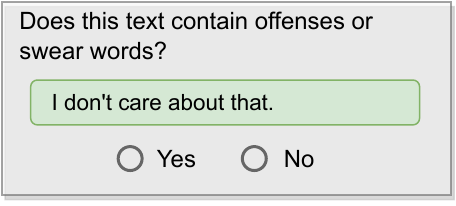}
\caption{Interface (translated into English for illustration) of the toxicity classification task for data collection with crowdsourcing.}
\label{fig:toxicity_task}
\end{figure}

\subsection{Semi-synthetic Dataset with Toxic Keywords Filtering}
To obtain toxic samples for these approach, we filtered Ukrainian tweets corpus from~\cite{bobrovnik2019twt} based on toxic keywords~\cite{tox_uk_lexicon}. We provide the full description of toxic keywords list construction in Appendix~\ref{sec:toxicity_keywords}. Then, tweets that did not contain any toxic words and additional texts from news and fiction UD Ukrainian IU dataset~\cite{udiu2016} were considered as non-toxic.

\subsection{Data Filtering with Crowdsourcing}

To obtain toxic samples with crowdsourcing, we took Ukrainian tweets corpus~\cite{bobrovnik2019twt}, erased URL links, and Twitter nicknames, dropped phrases with less than five and more than twenty words, randomly sampled texts for the annotation with Toloka platform\footnote{\href{https://toloka.ai}{https://toloka.ai}} (Figure~\ref{fig:toxicity_task}). We hired only workers who passed the in-platform test of Ukrainian language knowledge. Each task page contained 9 real tasks, 2 control tasks with known answers, and 1 training task with known answers and explanations. We blocked participants if their answers were inadequately fast (less than 15 seconds per page), if they skipped 5 pages in a row, or if they failed on more than 60\% of tasks with known answers. The crowdsourcing instructions and interface are listed in Appendix~\ref{sec:crowdsource_instruct}.

\section{Experimental Setup}

The statistics of train/val/test splits are presented in Table~\ref{tab:dataset_stat}. For the Ukrainian texts encoder, XLM-RoBERTa\footnote{\href{https://huggingface.co/FacebookAI/xlm-roberta-large}{https://huggingface.co/FacebookAI/xlm-roberta-large}}~\cite{conneau2019unsupervised} has already been proven as a strong baseline for multiple languages~\cite{imanigooghari2023glot500}. For LLMs prompting, we experimented with couple setups choosing LLaMa-2\footnote{\href{https://huggingface.co/meta-llama/Llama-2-7b-chat-hf}{https://huggingface.co/meta-llama/Llama-2-7b-chat-hf}}~\cite{touvron2023llama} and Mistral\footnote{\href{https://huggingface.co/mistralai/Mistral-7B-v0.1}{https://huggingface.co/mistralai/Mistral-7B-v0.1}}~\cite{DBLP:journals/corr/abs-2310-06825} as the most promising models for the Ukrainian inputs processing. For English toxicity classifier, we used an open fine-tuned version of the DistilBERT model to classify toxic comments.\footnote{\href{https://huggingface.co/martin-ha/toxic-comment-model}{https://huggingface.co/martin-ha/toxic-comment-model}}

\section{Results}
The classification results are presented in Table~\ref{tab:all_results}. 
% We tested the cross-lingual approaches and fine-tuned XLM-RoBERTa on the acquired datasets.
% 
Within methods that do not require fine-tuning, Backtranslation and Adapter Training look like promising baselines. Mistral outperforms LLaMa with top results on the semi-synthetic test set, but poorly on translated and, most importantly, crowdsourced data. At the same time, Backtranslation achieved top results on these two datasets that illustrates real Ukrainian toxic data the most.

When fine-tuned on the crowdsourced data, XLM-R exhibits almost perfect performance on both the in- and out-of-domain test sets. Undoubtedly, data collected through human annotations embodies the most accurate understanding of toxicity. However, its performance significantly drops on translated data with the results even lower than unsupervised approaches. That can be due to the reduced toxicity in the translated data: not all labelled originally toxic data remained toxic in Ukrainian. Conversely, the model fine-tuned on the translated data demonstrates the best results on the annotated test set. Thus, the Training Data Translation approach still stands as a viable baseline, showcasing robustness across out-of-domain data.

\section{Conclusion}
We presented the first of its kind study in toxicity detection in the Ukrainian language. Firstly, we tested several cross-lingual knowledge transfer approaches for the task that have different resources requirements: Backtranslation that requires three inferences steps, LLMs prompting, and Adapter training that requires only adapter layer fine-tuning. Still, the Backtranslation approach showed the best performance within unsupervised baselines.

Next, we explored three methods for acquiring a binary toxicity classification corpus: translating an existing labeled English dataset, filtering toxic samples using a predefined list of Ukrainian toxic keywords, and collecting data through crowdsourcing. The model fine-tuned on translated data exhibited the most resilient performance across out-of-domain datasets, serving as a robust baseline. Ultimately, the model fine-tuned on manually annotated data demonstrated the highest performance.

\section*{Limitations \& Ethics Statement}
In this work, we encounter toxic speech as only speech with obscene lexicon and commonly referred to as vulgar or profane language~\cite{costa2022no}. Thus, this work does not cover any other sides and shades of offensive language like hate, sarcasm, racism, sexism, etc. We believe that this study in toxic language detection will build a new foundation of any harmful language detection in Ukrainian.

Another limitation of this work that we consider only resource-rich language as English. For translated corpus acquisition it might also be beneficial to explore other languages from the linguistic families that are closer to Ukrainian, i.e. Polish or Croatian, if the corpora for the desired task exist in the corresponding languages.

In conclusion, the proposed toxicity detection model is openly shared with the community for further exploration. Deploying this model for specific use cases and domains should be complemented by human-computer interaction solutions that uphold users' freedom of speech while fostering proactive conversations. We firmly believe that our proposed toxicity classification data and models will contribute to the development of more fair and safe multilingual LLMs.

% \section*{Ethics Statement}

\bibliography{anthology,custom}

\begin{thebibliography}{29}
\expandafter\ifx\csname natexlab\endcsname\relax\def\natexlab#1{#1}\fi

\bibitem[{Aly et~al.(2023)Aly, Shi, Lin, Zhang, and Wilson}]{DBLP:conf/emnlp/AlySLZW23}
Rami Aly, Xingjian Shi, Kaixiang Lin, Aston Zhang, and Andrew~Gordon Wilson. 2023.
\newblock \href {https://aclanthology.org/2023.findings-emnlp.158} {Automated few-shot classification with instruction-finetuned language models}.
\newblock In \emph{Findings of the Association for Computational Linguistics: {EMNLP} 2023, Singapore, December 6-10, 2023}, pages 2414--2432. Association for Computational Linguistics.

\bibitem[{Bobrovnyk(2019{\natexlab{a}})}]{bobrovnik2019twt}
Kateryna Bobrovnyk. 2019{\natexlab{a}}.
\newblock Automated building and analysis of ukrainian twitter corpus for toxic text detection.
\newblock In \emph{COLINS 2019. Volume II: Workshop}.

\bibitem[{Bobrovnyk(2019{\natexlab{b}})}]{tox_uk_lexicon}
Kateryna Bobrovnyk. 2019{\natexlab{b}}.
\newblock Ukrainian obscene lexicon.
\newblock \href{https://github.com/saganoren/obscene-ukr}{https://github.com/saganoren/obscene-ukr}.
\newblock Accessed: 2023-12-14.

\bibitem[{Chung et~al.(2022)Chung, Hou, Longpre, Zoph, Tay, Fedus, Li, Wang, Dehghani, Brahma, Webson, Gu, Dai, Suzgun, Chen, Chowdhery, Narang, Mishra, Yu, Zhao, Huang, Dai, Yu, Petrov, Chi, Dean, Devlin, Roberts, Zhou, Le, and Wei}]{chung2022scaling}
Hyung~Won Chung, Le~Hou, Shayne Longpre, Barret Zoph, Yi~Tay, William Fedus, Eric Li, Xuezhi Wang, Mostafa Dehghani, Siddhartha Brahma, Albert Webson, Shixiang~Shane Gu, Zhuyun Dai, Mirac Suzgun, Xinyun Chen, Aakanksha Chowdhery, Sharan Narang, Gaurav Mishra, Adams Yu, Vincent~Y. Zhao, Yanping Huang, Andrew~M. Dai, Hongkun Yu, Slav Petrov, Ed~H. Chi, Jeff Dean, Jacob Devlin, Adam Roberts, Denny Zhou, Quoc~V. Le, and Jason Wei. 2022.
\newblock \href {https://doi.org/10.48550/ARXIV.2210.11416} {Scaling instruction-finetuned language models}.
\newblock \emph{CoRR}, abs/2210.11416.

\bibitem[{Conneau et~al.(2020)Conneau, Khandelwal, Goyal, Chaudhary, Wenzek, Guzm{\'{a}}n, Grave, Ott, Zettlemoyer, and Stoyanov}]{conneau2019unsupervised}
Alexis Conneau, Kartikay Khandelwal, Naman Goyal, Vishrav Chaudhary, Guillaume Wenzek, Francisco Guzm{\'{a}}n, Edouard Grave, Myle Ott, Luke Zettlemoyer, and Veselin Stoyanov. 2020.
\newblock \href {https://doi.org/10.18653/V1/2020.ACL-MAIN.747} {Unsupervised cross-lingual representation learning at scale}.
\newblock In \emph{Proceedings of the 58th Annual Meeting of the Association for Computational Linguistics, {ACL} 2020, Online, July 5-10, 2020}, pages 8440--8451. Association for Computational Linguistics.

\bibitem[{Costa{-}juss{\`{a}} et~al.(2022)Costa{-}juss{\`{a}}, Cross, {\c{C}}elebi, Elbayad, Heafield, Heffernan, Kalbassi, Lam, Licht, Maillard, Sun, Wang, Wenzek, Youngblood, Akula, Barrault, Gonzalez, Hansanti, Hoffman, Jarrett, Sadagopan, Rowe, Spruit, Tran, Andrews, Ayan, Bhosale, Edunov, Fan, Gao, Goswami, Guzm{\'{a}}n, Koehn, Mourachko, Ropers, Saleem, Schwenk, and Wang}]{costa2022no}
Marta~R. Costa{-}juss{\`{a}}, James Cross, Onur {\c{C}}elebi, Maha Elbayad, Kenneth Heafield, Kevin Heffernan, Elahe Kalbassi, Janice Lam, Daniel Licht, Jean Maillard, Anna~Y. Sun, Skyler Wang, Guillaume Wenzek, Al~Youngblood, Bapi Akula, Lo{\"{\i}}c Barrault, Gabriel~Mejia Gonzalez, Prangthip Hansanti, John Hoffman, Semarley Jarrett, Kaushik~Ram Sadagopan, Dirk Rowe, Shannon Spruit, Chau Tran, Pierre Andrews, Necip~Fazil Ayan, Shruti Bhosale, Sergey Edunov, Angela Fan, Cynthia Gao, Vedanuj Goswami, Francisco Guzm{\'{a}}n, Philipp Koehn, Alexandre Mourachko, Christophe Ropers, Safiyyah Saleem, Holger Schwenk, and Jeff Wang. 2022.
\newblock \href {https://doi.org/10.48550/ARXIV.2207.04672} {No language left behind: Scaling human-centered machine translation}.
\newblock \emph{CoRR}, abs/2207.04672.

\bibitem[{Das et~al.(2023)Das, Pandey, and Mukherjee}]{das2023evaluating}
Mithun Das, Saurabh~Kumar Pandey, and Animesh Mukherjee. 2023.
\newblock \href {https://doi.org/10.48550/ARXIV.2305.13276} {Evaluating chatgpt's performance for multilingual and emoji-based hate speech detection}.
\newblock \emph{CoRR}, abs/2305.13276.

\bibitem[{Devlin et~al.(2019)Devlin, Chang, Lee, and Toutanova}]{devlin-etal-2019-bert}
Jacob Devlin, Ming-Wei Chang, Kenton Lee, and Kristina Toutanova. 2019.
\newblock \href {https://doi.org/10.18653/v1/N19-1423} {{BERT}: Pre-training of deep bidirectional transformers for language understanding}.
\newblock In \emph{Proceedings of the 2019 Conference of the North {A}merican Chapter of the Association for Computational Linguistics: Human Language Technologies, Volume 1 (Long and Short Papers)}, pages 4171--4186, Minneapolis, Minnesota. Association for Computational Linguistics.

\bibitem[{El{-}Alami et~al.(2022)El{-}Alami, Alaoui, and En{-}Nahnahi}]{el2022multilingual}
Fatima{-}Zahra El{-}Alami, Said Ouatik~El Alaoui, and Noureddine En{-}Nahnahi. 2022.
\newblock \href {https://doi.org/10.1016/J.JKSUCI.2021.07.013} {A multilingual offensive language detection method based on transfer learning from transformer fine-tuning model}.
\newblock \emph{J. King Saud Univ. Comput. Inf. Sci.}, 34(8 Part {B}):6048--6056.

\bibitem[{Gilda et~al.(2021)Gilda, Giovanini, Silva, and Oliveira}]{DBLP:conf/euspn/GildaGS021}
Shlok Gilda, Luiz Giovanini, Mirela Silva, and Daniela Oliveira. 2021.
\newblock \href {https://doi.org/10.1016/J.PROCS.2021.12.254} {Predicting different types of subtle toxicity in unhealthy online conversations}.
\newblock In \emph{The 12th International Conference on Emerging Ubiquitous Systems and Pervasive Networks {(EUSPN} 2021) / The 11th International Conference on Current and Future Trends of Information and Communication Technologies in Healthcare (ICTH-2021), Leuven, Belgium, November 1-4, 2021}, volume 198 of \emph{Procedia Computer Science}, pages 360--366. Elsevier.

\bibitem[{ImaniGooghari et~al.(2023)ImaniGooghari, Lin, Kargaran, Severini, Sabet, Kassner, Ma, Schmid, Martins, Yvon et~al.}]{imanigooghari2023glot500}
Ayyoob ImaniGooghari, Peiqin Lin, Amir~Hossein Kargaran, Silvia Severini, Masoud~Jalili Sabet, Nora Kassner, Chunlan Ma, Helmut Schmid, Andr{\'e}~FT Martins, Fran{\c{c}}ois Yvon, et~al. 2023.
\newblock Glot500: Scaling multilingual corpora and language models to 500 languages.
\newblock \emph{arXiv preprint arXiv:2305.12182}.

\bibitem[{Jiang et~al.(2023)Jiang, Sablayrolles, Mensch, Bamford, Chaplot, de~Las~Casas, Bressand, Lengyel, Lample, Saulnier, Lavaud, Lachaux, Stock, Scao, Lavril, Wang, Lacroix, and Sayed}]{DBLP:journals/corr/abs-2310-06825}
Albert~Q. Jiang, Alexandre Sablayrolles, Arthur Mensch, Chris Bamford, Devendra~Singh Chaplot, Diego de~Las~Casas, Florian Bressand, Gianna Lengyel, Guillaume Lample, Lucile Saulnier, L{\'{e}}lio~Renard Lavaud, Marie{-}Anne Lachaux, Pierre Stock, Teven~Le Scao, Thibaut Lavril, Thomas Wang, Timoth{\'{e}}e Lacroix, and William~El Sayed. 2023.
\newblock \href {https://doi.org/10.48550/ARXIV.2310.06825} {Mistral 7b}.
\newblock \emph{CoRR}, abs/2310.06825.

\bibitem[{Jigsaw(2017)}]{jigsaw}
Jigsaw. 2017.
\newblock Toxic comment classification challenge.
\newblock \href{https://www.kaggle.com/c/jigsaw-toxic-comment-classification-challenge}{https://www.kaggle.com/c/jigsaw-toxic-comment-classification-challenge}.
\newblock Accessed: 2024-03-18.

\bibitem[{Jigsaw(2020)}]{jigsaw_multilingual}
Jigsaw. 2020.
\newblock Multilingual toxic comment classification.
\newblock \href{https://www.kaggle.com/c/jigsaw-multilingual-toxic-comment-classification}{https://www.kaggle.com/c/jigsaw-multilingual-toxic-comment-classification}.
\newblock Accessed: 2024-03-18.

\bibitem[{Jokic et~al.(2021)Jokic, Stankovic, Krstev, and Sandrih}]{DBLP:conf/ldk/JokicSKS21}
Danka Jokic, Ranka Stankovic, Cvetana Krstev, and Branislava Sandrih. 2021.
\newblock \href {https://doi.org/10.4230/OASICS.LDK.2021.13} {A twitter corpus and lexicon for abusive speech detection in serbian}.
\newblock In \emph{3rd Conference on Language, Data and Knowledge, {LDK} 2021, September 1-3, 2021, Zaragoza, Spain}, volume~93 of \emph{OASIcs}, pages 13:1--13:17. Schloss Dagstuhl - Leibniz-Zentrum f{\"{u}}r Informatik.

\bibitem[{Kotsyba et~al.(2016)Kotsyba, Moskalevskyi, Romanenko, Samoridna, Kosovska, Lytvyn, and Orlenko}]{udiu2016}
Natalia Kotsyba, Bohdan Moskalevskyi, Mykhailo Romanenko, Halyna Samoridna, Ivanka Kosovska, Olha Lytvyn, and Oksana Orlenko. 2016.
\newblock Ud ukrainian iu.
\newblock \url{ https://universaldependencies.org/treebanks/uk_iu/index.html}.

\bibitem[{Kumar et~al.(2023)Kumar, Pathania, and Raman}]{DBLP:journals/apin/KumarPR23}
Puneet Kumar, Kshitij Pathania, and Balasubramanian Raman. 2023.
\newblock \href {https://doi.org/10.1007/S10489-022-04046-6} {Zero-shot learning based cross-lingual sentiment analysis for sanskrit text with insufficient labeled data}.
\newblock \emph{Appl. Intell.}, 53(9):10096--10113.

\bibitem[{Logacheva et~al.(2022)Logacheva, Dementieva, Ustyantsev, Moskovskiy, Dale, Krotova, Semenov, and Panchenko}]{logacheva-etal-2022-paradetox}
Varvara Logacheva, Daryna Dementieva, Sergey Ustyantsev, Daniil Moskovskiy, David Dale, Irina Krotova, Nikita Semenov, and Alexander Panchenko. 2022.
\newblock \href {https://doi.org/10.18653/v1/2022.acl-long.469} {{P}ara{D}etox: Detoxification with parallel data}.
\newblock In \emph{Proceedings of the 60th Annual Meeting of the Association for Computational Linguistics (Volume 1: Long Papers)}, pages 6804--6818, Dublin, Ireland. Association for Computational Linguistics.

\bibitem[{Muennighoff et~al.(2023)Muennighoff, Wang, Sutawika, Roberts, Biderman, Scao, Bari, Shen, Yong, Schoelkopf, Tang, Radev, Aji, Almubarak, Albanie, Alyafeai, Webson, Raff, and Raffel}]{muennighoff2022crosslingual}
Niklas Muennighoff, Thomas Wang, Lintang Sutawika, Adam Roberts, Stella Biderman, Teven~Le Scao, M.~Saiful Bari, Sheng Shen, Zheng~Xin Yong, Hailey Schoelkopf, Xiangru Tang, Dragomir Radev, Alham~Fikri Aji, Khalid Almubarak, Samuel Albanie, Zaid Alyafeai, Albert Webson, Edward Raff, and Colin Raffel. 2023.
\newblock \href {https://doi.org/10.18653/V1/2023.ACL-LONG.891} {Crosslingual generalization through multitask finetuning}.
\newblock In \emph{Proceedings of the 61st Annual Meeting of the Association for Computational Linguistics (Volume 1: Long Papers), {ACL} 2023, Toronto, Canada, July 9-14, 2023}, pages 15991--16111. Association for Computational Linguistics.

\bibitem[{Oliinyk and Matviichuk(2023)}]{oliinyk2023low}
V~Oliinyk and I~Matviichuk. 2023.
\newblock Low-resource text classification using cross-lingual models for bullying detection in the ukrainian language.
\newblock \emph{\cyr{Адаптивні системи автоматичного управління: міжвідомчий науково-технічний збірник, 2023,№ 1 (42)}}.

\bibitem[{Pfeiffer et~al.(2020)Pfeiffer, R{\"u}ckl{\'e}, Poth, Kamath, Vuli{\'c}, Ruder, Cho, and Gurevych}]{pfeiffer-etal-2020-adapterhub}
Jonas Pfeiffer, Andreas R{\"u}ckl{\'e}, Clifton Poth, Aishwarya Kamath, Ivan Vuli{\'c}, Sebastian Ruder, Kyunghyun Cho, and Iryna Gurevych. 2020.
\newblock \href {https://doi.org/10.18653/v1/2020.emnlp-demos.7} {{A}dapter{H}ub: A framework for adapting transformers}.
\newblock In \emph{Proceedings of the 2020 Conference on Empirical Methods in Natural Language Processing: System Demonstrations}, pages 46--54, Online. Association for Computational Linguistics.

\bibitem[{Price et~al.(2020)Price, Gifford{-}Moore, Flemming, Musker, Roichman, Sylvain, Thain, Dixon, and Sorensen}]{DBLP:conf/acl-alw/PriceGFMRSTDS20}
Ilan Price, Jordan Gifford{-}Moore, Jory Flemming, Saul Musker, Maayan Roichman, Guillaume Sylvain, Nithum Thain, Lucas Dixon, and Jeffrey Sorensen. 2020.
\newblock \href {https://doi.org/10.18653/V1/2020.ALW-1.15} {Six attributes of unhealthy conversations}.
\newblock In \emph{Proceedings of the Fourth Workshop on Online Abuse and Harms, {WOAH} 2020, Online, November 20, 2020}, pages 114--124. Association for Computational Linguistics.

\bibitem[{Ptaszynski et~al.(2024)Ptaszynski, Pieciukiewicz, Dybala, Skrzek, Soliwoda, Fortuna, Leliwa, and Wroczynski}]{DBLP:journals/data/PtaszynskiPDSSF24}
Michal Ptaszynski, Agata Pieciukiewicz, Pawel Dybala, Pawel Skrzek, Kamil Soliwoda, Marcin Fortuna, Gniewosz Leliwa, and Michal Wroczynski. 2024.
\newblock \href {https://doi.org/10.3390/DATA9010001} {Expert-annotated dataset to study cyberbullying in polish language}.
\newblock \emph{Data}, 9(1):1.

\bibitem[{Tang et~al.(2020)Tang, Tran, Li, Chen, Goyal, Chaudhary, Gu, and Fan}]{tang2020multilingual}
Yuqing Tang, Chau Tran, Xian Li, Peng{-}Jen Chen, Naman Goyal, Vishrav Chaudhary, Jiatao Gu, and Angela Fan. 2020.
\newblock \href {http://arxiv.org/abs/2008.00401} {Multilingual translation with extensible multilingual pretraining and finetuning}.
\newblock \emph{CoRR}, abs/2008.00401.

\bibitem[{Tiedemann(2012)}]{tiedemann-2012-parallel}
J{\"o}rg Tiedemann. 2012.
\newblock \href {http://www.lrec-conf.org/proceedings/lrec2012/pdf/463_Paper.pdf} {Parallel data, tools and interfaces in {OPUS}}.
\newblock In \emph{Proceedings of the Eighth International Conference on Language Resources and Evaluation ({LREC}'12)}, pages 2214--2218, Istanbul, Turkey. European Language Resources Association (ELRA).

\bibitem[{Touvron et~al.(2023)Touvron, Martin, Stone, Albert, Almahairi, Babaei, Bashlykov, Batra, Bhargava, Bhosale, Bikel, Blecher, Canton{-}Ferrer, Chen, Cucurull, Esiobu, Fernandes, Fu, Fu, Fuller, Gao, Goswami, Goyal, Hartshorn, Hosseini, Hou, Inan, Kardas, Kerkez, Khabsa, Kloumann, Korenev, Koura, Lachaux, Lavril, Lee, Liskovich, Lu, Mao, Martinet, Mihaylov, Mishra, Molybog, Nie, Poulton, Reizenstein, Rungta, Saladi, Schelten, Silva, Smith, Subramanian, Tan, Tang, Taylor, Williams, Kuan, Xu, Yan, Zarov, Zhang, Fan, Kambadur, Narang, Rodriguez, Stojnic, Edunov, and Scialom}]{touvron2023llama}
Hugo Touvron, Louis Martin, Kevin Stone, Peter Albert, Amjad Almahairi, Yasmine Babaei, Nikolay Bashlykov, Soumya Batra, Prajjwal Bhargava, Shruti Bhosale, Dan Bikel, Lukas Blecher, Cristian Canton{-}Ferrer, Moya Chen, Guillem Cucurull, David Esiobu, Jude Fernandes, Jeremy Fu, Wenyin Fu, Brian Fuller, Cynthia Gao, Vedanuj Goswami, Naman Goyal, Anthony Hartshorn, Saghar Hosseini, Rui Hou, Hakan Inan, Marcin Kardas, Viktor Kerkez, Madian Khabsa, Isabel Kloumann, Artem Korenev, Punit~Singh Koura, Marie{-}Anne Lachaux, Thibaut Lavril, Jenya Lee, Diana Liskovich, Yinghai Lu, Yuning Mao, Xavier Martinet, Todor Mihaylov, Pushkar Mishra, Igor Molybog, Yixin Nie, Andrew Poulton, Jeremy Reizenstein, Rashi Rungta, Kalyan Saladi, Alan Schelten, Ruan Silva, Eric~Michael Smith, Ranjan Subramanian, Xiaoqing~Ellen Tan, Binh Tang, Ross Taylor, Adina Williams, Jian~Xiang Kuan, Puxin Xu, Zheng Yan, Iliyan Zarov, Yuchen Zhang, Angela Fan, Melanie Kambadur, Sharan Narang, Aur{\'{e}}lien Rodriguez, Robert Stojnic, Sergey Edunov,
  and Thomas Scialom. 2023.
\newblock \href {https://doi.org/10.48550/ARXIV.2307.09288} {Llama 2: Open foundation and fine-tuned chat models}.
\newblock \emph{CoRR}, abs/2307.09288.

\bibitem[{Wadud et~al.(2023)Wadud, Mridha, Shin, Nur, and Saha}]{DBLP:journals/csse/WadudMSNS23}
Md. Anwar~Hussen Wadud, Muhammad~F. Mridha, Jungpil Shin, Kamruddin Nur, and Aloke~Kumar Saha. 2023.
\newblock \href {https://doi.org/10.32604/csse.2023.027841} {Deep-bert: Transfer learning for classifying multilingual offensive texts on social media}.
\newblock \emph{Comput. Syst. Sci. Eng.}, 44(2):1775--1791.

\bibitem[{Wei et~al.(2022)Wei, Tay, Bommasani, Raffel, Zoph, Borgeaud, Yogatama, Bosma, Zhou, Metzler, Chi, Hashimoto, Vinyals, Liang, Dean, and Fedus}]{wei2022emergent}
Jason Wei, Yi~Tay, Rishi Bommasani, Colin Raffel, Barret Zoph, Sebastian Borgeaud, Dani Yogatama, Maarten Bosma, Denny Zhou, Donald Metzler, Ed~H. Chi, Tatsunori Hashimoto, Oriol Vinyals, Percy Liang, Jeff Dean, and William Fedus. 2022.
\newblock \href {https://openreview.net/forum?id=yzkSU5zdwD} {Emergent abilities of large language models}.
\newblock \emph{Trans. Mach. Learn. Res.}, 2022.

\bibitem[{Xue et~al.(2021)Xue, Constant, Roberts, Kale, Al{-}Rfou, Siddhant, Barua, and Raffel}]{xue2020mt5}
Linting Xue, Noah Constant, Adam Roberts, Mihir Kale, Rami Al{-}Rfou, Aditya Siddhant, Aditya Barua, and Colin Raffel. 2021.
\newblock \href {https://doi.org/10.18653/V1/2021.NAACL-MAIN.41} {mt5: {A} massively multilingual pre-trained text-to-text transformer}.
\newblock In \emph{Proceedings of the 2021 Conference of the North American Chapter of the Association for Computational Linguistics: Human Language Technologies, {NAACL-HLT} 2021, Online, June 6-11, 2021}, pages 483--498. Association for Computational Linguistics.

\end{thebibliography}
\bibliographystyle{acl_natbib}

\onecolumn
\appendix

\section{The Full List of Toxic Keywords Used for Filtering}
\label{sec:toxicity_keywords}

\textbf{\textcolor{red}{\textit{This list only serves to increase reproducibility of our work and has no intention to offend the reader.}}}

Additionally to the openly available list of Ukrainian toxic keywords\footnote{\href{https://github.com/saganoren/obscene-ukr}{https://github.com/saganoren/obscene-ukr}}, we also came up with some additional words that can be divided into the following groups: 

Slurs towards a group of people under discrimination (nationality, race, sexual orientation etc.):

\textit{\foreignlanguage{ukrainian}{``хохол'', ``хохли'', ``хохлом'', ``хохлами'', ``жид'', ``жиди'', ``жидом'', ``жидами'', ``жидовка'', ``жидовський'', ``жидовські'', ``жидовська'', ``вузькоглазий'',  ``вузькоглазі'', ``ніга'', ``нігга'', ``ніггерам'', ``ніггери'', ``ніггерів'', ``нігер'', ``нігера'', ``нігерами'', ``нігери'', ``нігерка'', ``нігерська'', ``нігерський'', ``нігерських'', ``нігерські'', ``нігерів'', ``нігріла'', ``нігєр'', ``нігєра'', ``нігєри'', ``нігєрський'', ``педарастів'', ``педераст'', ``педерастія'', ``педик'', ``педики'', ``педиків'', ``педік'', ``педіковського'', ``підар'', ``підари'', ``підором'', ``підараси'', ``підорський'', ``підорас'', ``підарас'', ``підарам'', ``підару'', ``підарасу'', ``підарасам'', ``тьолка'', ``тьолкою'', ``тьолки'', ``тьолками'', ``тьолкам'', ``тьолці'', ``блядь'', ``бляді'', ``шалава'', ``шалави'', ``прошмандовка''}}

Most often used toxic or hate appeals to the opponent:

\foreignlanguage{ukrainian}{\textit{''уйобок'', ''хуйло'', ''ахуєл'', ''уєбан'', ``уїбан'', ``довбойоб'', ``долбойоб'', ``залупа'', ``гандон'', ``пизда'', ``їблан'', ``єблан'', ``їбанутий'', ``єбанутий''}}

Different obscene words (forms without endings):

\foreignlanguage{ukrainian}{\textit{``їбат'', ``їбан'', ``пизд'', ``бля''}}

\section{LLM Toxicity Classification Prompt}
\label{sec:app_llm_prompts}

Denote: even if we perform classification for texts in Ukrainian, the core structure of the note is still in Enlglish. Such a design was proven to be the most successful in our experiments.

\textit{\newline Classify the text into two categories: contains obscene words or contains none obscene words. Reply with only one word: obscene or normal.
\newline
\newline
Examples:
\newline
Text: \foreignlanguage{ukrainian}{Сьогодні знайти у відкритих джерелах точну суму, витрачену на будівництво об’єкта, що про нього мова, майже неможливо.}
Sentiment: normal.
\newline
Text: \foreignlanguage{ukrainian}{знаєте, якщо свої дебільні коментарі ще й ілюструвати посиланнями на російську вікі, хтось може здогадатися, що ви тупий єблан.}
Sentiment: obscene.
\newline
\newline
Text: \{text\}
\newline
Sentiment:
}

\newpage

\section{Corpora Data Examples}
\label{sec:app_data}

\subsection{Translated Data}
\label{sec:app_translated_data}

Here, we provide examples in Table~\ref{tab:app_toxicity_translated_examples} of data gained from the Data Translation approach. The translation are the ones obtained from the utilized translation model~\cite{tiedemann-2012-parallel}.

\begin{table}[h!]
    \centering
    \footnotesize
    \begin{tabular}{p{2cm}|p{11cm}}
    \toprule
        
        % \multicolumn{2}{c}{\textbf{Toxicity Classification}} \\
        % \midrule 
        Toxic & \foreignlanguage{ukrainian}{ви всі тупі осли.} \newline \textcolor{gray}{\scriptsize{\textit{youre all dumb asses}}} \\ 
        \midrule 
        Non-toxic & \foreignlanguage{ukrainian}{Є два адміністратори, які досить добре працюють з такими статтями, можливо, ви могли б зв'язатися з ними.} \newline \textcolor{gray}{\scriptsize{\textit{there are two admins that do handle such articles pretty well you could maybe contact  and }}} \\
        \midrule 
        Toxic & \foreignlanguage{ukrainian}{І Роберт - це чорне лайно} \newline \textcolor{gray}{\scriptsize{\textit{and robert is a black shit}}} \\ 
        \midrule 
        Non-toxic & \foreignlanguage{ukrainian}{Гаразд, я почав трансляцію нової статті, я використав вашу запропоновану назву може використовувати більше деталей зараз} \newline \textcolor{gray}{\scriptsize{\textit{ok i started the transtion the new article i used your suggested title could use a lot more detail now}}} \\
        \midrule 
        Toxic & \foreignlanguage{ukrainian}{Що за купа ср*них ботанів?} \newline \textcolor{gray}{\scriptsize{\textit{what a bunch of f**king nerds}}} \\ 
        \midrule 
        Non-toxic & \foreignlanguage{ukrainian}{Зупиніться, будь ласка, якщо ви продовжите вандализувати сторінки, ви будете заблоковані від редагування wikipedia} \newline \textcolor{gray}{\scriptsize{\textit{please stop if you continue to vandalize wikipedia you will be blocked from editing}}} \\
        \midrule 
        Toxic & \foreignlanguage{ukrainian}{Альтернативна поп-культура, що означає п*зда чи ци, розкидає таємничу сучку, яка руйнує все, що примара називає когось, це спосіб дати людині знати, що вони є п*зда в той час як цензують інших навколо вас в громадських місцях або в соціальних кутах, сучасний сленг попереджаючи інших про небезпеку.} \newline \textcolor{gray}{\scriptsize{\textit{alternative  pop culture meaning   c*nt or cee unt  a percieved mysterious bitch that destroys everything  whem calling someone this is a way of letting anyone know they are a c*nt while censoring others around you in public or in social corners  a modern slang alerting other of the danger}}} \\ 
        \midrule 
        Non-toxic & \foreignlanguage{ukrainian}{Адміністратори виконують дії, що ґрунтуються на громадському консенсусі, вони не приймають односторонніх рішень далі, тому у зв'язку з цим редактори, які зосереджують свою увагу на виборах або канадалях, не мають можливості перенаправити кандидатів на партійні статті.} \newline \textcolor{gray}{\scriptsize{\textit{  admins execute actions based on community consensus  they do not make unilateral decisions further that afd did not have the involvement of editors who focus on ontario or canadawide elections so they were likely unfamiliar with the option of redirecting to party candidate articles}}} \\
        \bottomrule
    \end{tabular}
    \caption{Examples of translated samples for \textbf{Toxicity Classification} task. English translation are taken from the Jigsaw dataset~\cite{jigsaw}.}
    \label{tab:app_toxicity_translated_examples}
\end{table}

\newpage

\subsection{Semi-synthetic Data}
\label{sec:app_semi_synthetic_data}

Here, we provide examples in Table~\ref{tab:app_toxicity_semisynthetic_examples} of data gained by filtering with toxic keywords.

\begin{table}[h!]
    \centering
    \footnotesize
    \begin{tabular}{p{2cm}|p{11cm}}
    \toprule
        
        % \multicolumn{2}{c}{\textbf{Toxicity Classification}} \\
        % \midrule 
        Toxic & \foreignlanguage{ukrainian}{@USER нє, китай рулить, то однозначно. ден сяопін був генієм економіки. але це було підписано бо більше ні на шо пі**рович не заслужив:)} \newline \textcolor{gray}{\scriptsize{\textit{@USER no, the Chinese drive, of course. The shoopin was an economic genius. But it was signed because no more on the sublarcier was worthy of:)}}} \\ 
        \midrule 
        Non-toxic & \foreignlanguage{ukrainian}{@USER 1) доброго часу, коліжанці дав почитати збірку, багато що оцінила, але запитала про ''Самонедостатність'' ..} \newline \textcolor{gray}{\scriptsize{\textit{@USER 1) good time, the colts gave the collection a lot of reading and appreciated, but asked about ''Memonysity...''}}} \\
        \midrule 
        Toxic & \foreignlanguage{ukrainian}{вже не пі**рас?} \newline \textcolor{gray}{\scriptsize{\textit{Isn't that a f**got?}}} \\ 
        \midrule 
        Non-toxic & \foreignlanguage{ukrainian}{Не раз заявляв про наміри зайти на наш ринок ірландський Ryanair .} \newline \textcolor{gray}{\scriptsize{\textit{More than once, he claimed to visit our market in Irish Ryanair.}}} \\
        \midrule 
        Toxic & \foreignlanguage{ukrainian}{сьогоднішня мрія - адекватний транспорт в крим, щоб не доводилося щоразу мозок собі ї**ти стиковкою цих жахливих людиноненависницьких рейсів} \newline \textcolor{gray}{\scriptsize{\textit{Today's dream is a safe transport into the ice so that every brain doesn't have to f**k its way through these terrible man - hated flights.}}} \\ 
        \midrule 
        Non-toxic & \foreignlanguage{ukrainian}{Співрозмовники досягли домовленості про проведення чергового засідання Спільної міжурядової українсько - туркменської комісії з економічного та культурно - гуманітарного співробітництва вже ближчим часом .} \newline \textcolor{gray}{\scriptsize{\textit{Coordinators have reached an agreement to hold a joint Intergovernmental Union Commission on Economic and Cultural Cooperation for a longer time.}}} \\
        \midrule 
        Toxic & \foreignlanguage{ukrainian}{нема відчуття гіршого, ніж коли розумієш, шо ти конкретно так тупанув, і через це все йде по п**ді.} \newline \textcolor{gray}{\scriptsize{\textit{There's no worse feeling than when you realize that you were exactly f**king that way, and that's why everything goes on p*ss.}}} \\ 
        \midrule 
        Non-toxic & \foreignlanguage{ukrainian}{Державне підприємство « Конструкторське бюро „ Південне “ ім . М . К . Янгеля » було створено 1951 як конструкторський відділ Південного машинобудівного заводу з виробництва військових ракет .} \newline \textcolor{gray}{\scriptsize{\textit{The state enterprise (C) was created by 1951 as the South Carworker's design department for the production of military rockets.}}} \\
        \bottomrule
    \end{tabular}
    \caption{Examples of natural samples for \textbf{Toxicity Classification} task obtained from Ukrainian tweets corpus from~\cite{bobrovnik2019twt} and news and fiction UD Ukrainian IU dataset~\cite{udiu2016} by the toxic keywords filtering.}
    \label{tab:app_toxicity_semisynthetic_examples}
\end{table}

\newpage

\subsection{Crowdsourced Data}
\label{sec:app_semi_synthetic_data}

Here, we provide examples in Table~\ref{tab:app_toxicity_crowdsourced_examples} of data gained by crowdsource collection.

\begin{table}[h!]
    \centering
    \footnotesize
    \begin{tabular}{p{2cm}|p{11cm}}
    \toprule
        
        % \multicolumn{2}{c}{\textbf{Toxicity Classification}} \\
        % \midrule 
        Toxic & \foreignlanguage{ukrainian}{Фу ти що той як його п*дар?} \newline \textcolor{gray}{\scriptsize{\textit{What are you, like his f*got?}}} \\ 
        \midrule 
        Non-toxic & \foreignlanguage{ukrainian}{Якщо не думати про певні проблеми, то жити дуже навіть можна.} \newline \textcolor{gray}{\scriptsize{\textit{If you don't think about certain problems, you can live very well.}}} \\
        \midrule
        Toxic & \foreignlanguage{ukrainian}{П*здєц, я такий безкоштовно могла будь-де знайти} \newline \textcolor{gray}{\scriptsize{\textit{F*ck, I could have found one for free anywhere)}}} \\ 
        \midrule 
        Non-toxic & \foreignlanguage{ukrainian}{Бо я щось зовсім від новин відвик...} \newline \textcolor{gray}{\scriptsize{\textit{Because I'm not used to the news at all...}}} \\
        \midrule
        Toxic & \foreignlanguage{ukrainian}{Завод по проізводству сосатєльних х*ов.} \newline \textcolor{gray}{\scriptsize{\textit{A factory for the production of sucking d*ks.}}} \\ 
        \midrule 
        Non-toxic & \foreignlanguage{ukrainian}{Нарізав вам фрагменти вчорашнього ефіру з Мураєвим.} \newline \textcolor{gray}{\scriptsize{\textit{I've cut you fragments of yesterday's broadcast with Muraev.}}} \\
        
        \midrule
        Toxic & \foreignlanguage{ukrainian}{Тому от вони, а не х*рь якась} \newline \textcolor{gray}{\scriptsize{\textit{So here they are, not some bul*hit}}} \\ 
        \midrule 
        Non-toxic & \foreignlanguage{ukrainian}{Особливо, коли в тебе другий день шалена слабкість.} \newline \textcolor{gray}{\scriptsize{\textit{Especially when you've been feeling crazy weak for two days.}}} \\

        \midrule
        Toxic & \foreignlanguage{ukrainian}{Давайте, розкажіть нам що це просте співпадіння, оце х*та з Мо*нкою.} \newline \textcolor{gray}{\scriptsize{\textit{Go ahead, tell us that it's a simple coincidence, this f*k with the sc*tum.}}} \\ 
        \midrule 
        Non-toxic & \foreignlanguage{ukrainian}{Не люблю свята, бо це лише витрати та клопіт, а так жодної різниці зі звичайним рутинним днем.} \newline \textcolor{gray}{\scriptsize{\textit{I don't like holidays because they're just expenses and hassle, and there's no difference between them and a normal day.}}} \\

        \midrule
        Toxic & \foreignlanguage{ukrainian}{Ну для мене люба френдзона це ху*ве місце} \newline \textcolor{gray}{\scriptsize{\textit{Well, for me, any friendzone is a fu*ing place}}} \\ 
        \midrule 
        Non-toxic & \foreignlanguage{ukrainian}{Є цікаві персонажі й діалоги, сюжет середній.} \newline \textcolor{gray}{\scriptsize{\textit{There are interesting characters and dialogues, but the plot is average.}}} \\

        \bottomrule
    \end{tabular}
    \caption{Examples of crowdsourced samples for \textbf{Toxicity Classification} task obtained from Ukrainian tweets corpus from~\cite{bobrovnik2019twt}.}
    \label{tab:app_toxicity_crowdsourced_examples}
\end{table}

\newpage

\section{Crowdsourcing platform instructions and interface}
\label{sec:crowdsource_instruct}

Here, we list the full instruction and task interface in the original Ukrainian language. Per each page, the annotators were paid 0.10\$.

\subsection{General instructions for the task.}

\foreignlanguage{ukrainian}{Вам потрібно прочитати речення і визначити, чи містять вони образи або нецензурні та грубі слова.\newline \textcolor{gray}{\scriptsize{\textit{You need to read the sentences and determine if they contain insults or obscene and rude words.}}}

Увага! Необразне речення може містити критику і бути негативно забарвленим. \newline \textcolor{gray}{\scriptsize{\textit{WARNING! A non-figurative sentence can contain criticism and be negatively colored.}}}

Приклади \newline \textcolor{gray}{\scriptsize{\textit{Examples}}}

Образливі речення: \newline \textcolor{gray}{\scriptsize{\textit{Offensive sentences:}}}

    \begin{itemize}
        \item Інтернет-шпана, не тобі мене повчати.
        \newline \textcolor{gray}{\scriptsize{\textit{Internet-nasty crew, it's not for you to teach me.}}}
        \item Яка підписка, що ти несеш, поїхавший?
        \newline \textcolor{gray}{\scriptsize{\textit{What is the subscription, what are you talking about, are you mad?}}}
        \item Щонайменше два малолітніх дегенерати в треді, мда.
        \newline \textcolor{gray}{\scriptsize{\textit{At least two juvenile degenerates in a thread, huh?}}}
        \item Взагалі не бачу сенсу сперечатися з приводу дюймів, хуєвий там ips чи ні, машина не цим цікава.
        \newline \textcolor{gray}{\scriptsize{\textit{In general, I don't see any point in arguing about inches, whether the ips is fucked up or not, this is not what makes the car interesting.}}}
    \end{itemize}

Нейтральні (не образливі) речення:
\newline \textcolor{gray}{\scriptsize{\textit{Neutral (not offensive) sentences:}}}

    \begin{itemize}
        \item У нас є убунти і текнікал прев'ю.
        \newline \textcolor{gray}{\scriptsize{\textit{ We have Ubuntu and Teknical previews.}}}
        \item він теж був хоробрим!
        \newline \textcolor{gray}{\scriptsize{\textit{He was brave too!}}}
        \item Це безглуздо, ти ж знаєш
        \newline \textcolor{gray}{\scriptsize{\textit{It makes no sense, you know that.}}}
        \item Якщо він мріє напакостити своїм сусідам, то це погано.
        \newline \textcolor{gray}{\scriptsize{\textit{If he dreams of hurting his neighbors, that's bad.}}}

    \end{itemize}
    }

\subsection{Task interface}

\foreignlanguage{ukrainian}{Чи містить цей текст образи або нецензурні слова?}
\newline \textcolor{gray}{\scriptsize{Does the text contain insults or obscenities?}}
\begin{itemize}
    \item \foreignlanguage{ukrainian}{Так} 
    \newline \textcolor{gray}{\scriptsize{Yes}}
    \item \foreignlanguage{ukrainian}{Ні} 
    \newline \textcolor{gray}{\scriptsize{No}}
\end{itemize}

\end{document}